\documentclass[conference]{IEEEtran}
\IEEEoverridecommandlockouts
\makeatletter
\newcommand{\linebreakand}{\end{@IEEEauthorhalign}\hfill\mbox{}\par
	\mbox{}\hfill\begin{@IEEEauthorhalign}}
\makeatother
\usepackage{cite}
\usepackage{amsmath,amssymb,amsfonts}
\usepackage{algorithmic}
\usepackage{graphicx}
\usepackage{tikz}
\usetikzlibrary{arrows.meta,calc}
\usepackage{textcomp}
\usepackage{xcolor}
\usepackage{stfloats}
\usepackage[utf8]{inputenc}
\usepackage[T1]{fontenc}
\usepackage{url}
\usepackage{booktabs}
\usepackage{nicefrac}
\usepackage{microtype}
\usepackage{caption}
\usepackage{subcaption}
\usepackage{enumitem}
\usepackage{amsthm}
\usepackage{dsfont}
\usepackage{comment}
\usepackage{nameref}
\usepackage{soul}
\usepackage{transparent}

\DeclareMathOperator*{\argmin}{arg\,min}

\theoremstyle{plain}
\newtheorem{theorem}{Theorem}

\theoremstyle{definition}
\newtheorem{definition}[theorem]{Definition}

\theoremstyle{remark}

\def\BibTeX{{\rm B\kern-.05em{\sc i\kern-.025em b}\kern-.08em
    T\kern-.1667em\lower.7ex\hbox{E}\kern-.125emX}}
\begin{document}

\title{Integrating Fairness and Model Pruning \\ Through Bi-level Optimization}

\author{
	\IEEEauthorblockN{Yucong Dai}
	\IEEEauthorblockA{Electrical and Computer Engineering\\
		Clemson University\\
		Clemson, USA\\
		yucongd@clemson.edu}
	\and
	\IEEEauthorblockN{Gen Li}
	\IEEEauthorblockA{Department of Computer Science\\
		University of Houston\\
		Houston, TX, USA\\
		gli25@central.uh.edu}
	\and
	\IEEEauthorblockN{Feng Luo}
	\IEEEauthorblockA{School of Computing\\
		Clemson University\\
		Clemson, USA\\
		luofeng@clemson.edu}
	\linebreakand
	\IEEEauthorblockN{Xiaolong Ma}
	\IEEEauthorblockA{School of Electrical, Computing, and Software Engineering\\
		University of Arizona\\
		Tucson, AZ, USA\\
		xiaolongma@arizona.edu}
	\and
	\IEEEauthorblockN{Yongkai Wu}
	\IEEEauthorblockA{Electrical and Computer Engineering\\
		Clemson University\\
		Clemson, USA\\
		yongkaw@clemson.edu}
}

\maketitle

\begin{abstract}
	Deep neural networks achieve high predictive performance but often require substantial computational and storage resources, which motivates model compression techniques such as neural network pruning.
	Conventional pruning methods, however, can introduce or amplify algorithmic bias and produce disparate outcomes across demographic groups.
	We study fair model pruning, which seeks sparse models that satisfy fairness criteria while maintaining predictive performance.
	We characterize fairness using both group accuracy disparity and differences in compression-induced accuracy degradation across demographic groups.
	We propose a framework that jointly optimizes pruning masks and model weights under fairness constraints, thereby integrating accuracy and fairness into the compression process.
	We formulate fair pruning as a constrained bi-level optimization problem and derive an efficient optimization procedure.
	Experiments across multiple datasets, predictive tasks, model architectures, and sparsity levels show that our framework improves the trade-off among fairness, accuracy, and training efficiency relative to the baseline pruning strategies.
	Compared with the strongest evaluated baselines, BiFP maintains competitive accuracy and fairness while requiring more than $94\%$ fewer training iterations in the real-world experiments.
\end{abstract}


\section{Introduction}

Artificial intelligence applications across many domains rely largely on the predictive performance of deep neural networks (DNNs)~\cite{DBLP:conf/cvpr/HeZRS16,dong2015image}.
DNNs typically require many parameters (weights) to learn complex relationships in data, resulting in substantial computational and storage costs.
Various model compression techniques have therefore been developed to improve efficiency while preserving predictive performance~\cite{han2015deep,he2017channel,guo2016dynamic,molchanov2016pruning}.
However, model compression techniques such as neural network pruning can introduce or exacerbate societal biases~\cite{hooker2019compressed,liebenwein2021lost,DBLP:journals/corr/abs-2304-12622}.
These effects must be addressed before compressed models can be deployed reliably in real-world applications.

\begin{figure}[t]
	\includegraphics[width=\linewidth]{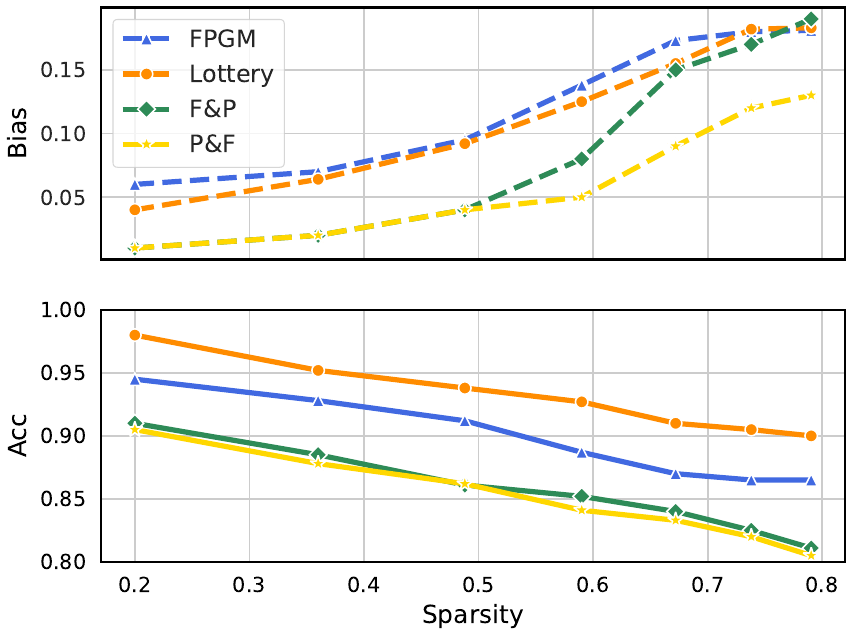}
	\caption{Accuracy and fairness disparity of conventional pruning methods and two sequential fairness-pruning strategies.}
	\label{fig:pruning-fairness-motivation}
\end{figure}

Mitigating demographic disparities in machine learning has received considerable attention~\cite{DBLP:conf/iclr/SalvadorCVMO22,DBLP:conf/aaai/WangW0L23,DBLP:conf/aaai/ZhuGWLHZ23}.
Common strategies include fairness-constrained optimization~\cite{DBLP:conf/aistats/ZafarVGG17} and adversarial learning~\cite{DBLP:conf/icml/MadrasCPZ18}.
Most existing methods focus on \textbf{dense} deep neural networks with large parameter spaces.
With a growing need for sparse neural networks, fair pruning must balance performance, fairness, and efficiency.
Achieving this balance is challenging as pruning identifies a weight mask while attempting to preserve accuracy, whereas conventional fairness methods adjust model weights to reduce bias.
The interaction among masks, weights, and fairness constraints makes it difficult to optimize accuracy and fairness simultaneously.

\begin{figure}[t]
	\centering
	\resizebox{\linewidth}{!}{\begin{tikzpicture}[x=.67cm,y=.92cm,
				font=\fontsize{8}{9.5}\selectfont,
				every node/.style={inner sep=0pt},
				state/.style={rounded corners=2pt,line width=.6pt,
						minimum width=.88cm,minimum height=.66cm,
						font=\fontsize{15}{17}\selectfont},
				weightedge/.style={-{Stealth[length=1.7mm,width=1.2mm]},line width=.75pt,draw=black},
				maskedge/.style={-{Stealth[length=1.7mm,width=1.2mm]},line width=.75pt,
						dash pattern=on .6pt off 2.2pt,line cap=round,draw=black},
				stepbadge/.style={circle,draw=black,fill=white,line width=.4pt,
						minimum size=.32cm,font=\fontsize{7}{8}\selectfont},
				note/.style={font=\fontsize{7}{8}\selectfont,text=black}]
			\definecolor{maskaccent}{RGB}{235,112,15}
			\definecolor{weightaccent}{RGB}{15,111,210}
			\path[use as bounding box] (0,-.55) rectangle (11.2,4.4);
			\fill[maskaccent!27] (0,2.36) rectangle (11.2,4.4);
			\fill[weightaccent!23] (0,.04) rectangle (11.2,2.36);

			\node[anchor=west,text=maskaccent,font=\bfseries\fontsize{8}{9.5}\selectfont]
			at (.10,4.15) {Upper level: fairness-aware mask learning};
			\draw[black,line width=.4pt] (.10,3.94)--(11.10,3.94);

			\foreach \i/\x in {0/1.10,1/3.80,2/6.50,3/9.20} {
					\node[state,draw=black,fill=maskaccent!65] (m\i) at (\x,3.16) {$m_{\i}$};
					\node[state,draw=black,fill=weightaccent!55] (w\i) at (\x,1.35) {$\theta_{\i}$};
				}
			\node[note] at (1.10,3.69) {Initial mask};
			\node[note,anchor=west] at (.10,.72) {Pretrained weights};

			\draw[weightedge,rounded corners=2pt]
			($(m0.east)+(0,-.13)$)--(1.93,3.03)--(1.93,2.36)
			--(.03,2.36)--(.03,1.35)--(w0.west);

			\foreach \prev/\next/\join/\right in {0/1/2.94/4.63,1/2/5.64/7.33,2/3/8.34/10.03} {
					\coordinate (maskjoin\next) at (\join,3.16);
					\draw[maskedge,-] (m\prev.east)--(maskjoin\next);
					\draw[maskedge,-] ($(w\prev.east)+(0,.10)$)
					.. controls ($(w\prev.east)+(1.00,.10)$)
					and ($(maskjoin\next)+(-.90,0)$) .. (maskjoin\next);
					\fill[black] (maskjoin\next) circle (.025);
					\draw[maskedge] (maskjoin\next)--(m\next.west);
					\coordinate (weightjoin\next) at ({\join-.25},1.35);
					\draw[weightedge,-] (w\prev.east)--(weightjoin\next);
					\draw[weightedge,-,rounded corners=2pt]
					($(m\next.east)+(0,-.13)$)--(\right,3.03)
					--(\right,2.36)--({\join-.25},2.36)--(weightjoin\next);
					\fill[black] (weightjoin\next) circle (.025);
					\draw[weightedge] (weightjoin\next)--(w\next.west);
				}
			\node[stepbadge] at (.55,2.04) {0};
			\node[stepbadge] at (2.65,3.49) {1};
			\node[stepbadge] at (2.65,1.12) {2};
			\node[stepbadge] at (5.35,3.49) {3};
			\node[stepbadge] at (5.35,1.12) {4};

			\node[stepbadge] at (8.05,3.49) {5};
			\node[stepbadge] at (8.05,1.12) {6};

			\draw[maskedge] (m3.east)--(10.35,3.16);
			\draw[weightedge] (w3.east)--(10.35,1.35);
			\node[font=\fontsize{13}{14}\selectfont,text=black] at (10.75,3.16) {$\cdots$};
			\node[font=\fontsize{13}{14}\selectfont,text=black] at (10.75,1.35) {$\cdots$};

			\draw[black,line width=.4pt] (.10,.53)--(11.10,.53);
			\node[anchor=west,text=weightaccent,font=\bfseries\fontsize{8}{9.5}\selectfont]
			at (.10,.29) {Lower level: fairness-aware weight learning};

			\draw[maskedge] (.35,-.25)--(.95,-.25);
			\node[anchor=west,note] at (1.10,-.25) {Mask update};
			\draw[weightedge] (4.35,-.25)--(4.95,-.25);
			\node[anchor=west,note] at (5.10,-.25) {Weight update};
			\node[stepbadge,minimum size=.23cm,font=\fontsize{6}{7}\selectfont] at (8.55,-.25) {$i$};
			\node[anchor=west,note] at (8.85,-.25) {Data flow};
		\end{tikzpicture}
	}
	\caption{Overview of Bi-level Fair Pruning (BiFP).
		Starting from an initial mask and pretrained weights, BiFP alternates between updating the mask from the previous mask and current weights (dotted arrows), and updating the weights from the previous weights and updated mask (solid arrows).
		Each pair of input paths joins before the target state.
		Step 0 denotes initialization; steps 1--6 illustrate three successive update rounds.
		Fairness is considered in both mask and weight learning.
	}
	\label{fig:bifp-framework}
\end{figure}

An intuitive approach is to apply fairness training and model pruning sequentially.
Pruning a fair model prioritizes a mask that maximizes accuracy, often overlooking fairness requirements.
Conversely, mitigating bias after pruning can reduce accuracy.
In Fig.~\ref{fig:pruning-fairness-motivation}, we compare four common pruning methods at various sparsity levels:
a) the classic structured pruning method (FPGM)~\cite{he2019filter},
b) the classic unstructured pruning method (Lottery)~\cite{frankle2018lottery},
c) building a fair model and subsequently pruning it (F\&P),
and d) pruning a model and then fine-tuning it with fairness constraints (P\&F).
The results show that pruning can introduce bias and that these sequential strategies do not consistently mitigate it.

These observations motivate a unified strategy that accounts for the interactions among masks, weights, and fairness constraints.
We present \underline{\textbf{Bi}}-level \underline{\textbf{F}}air \underline{\textbf{P}}runing (\textbf{BiFP}), which uses bi-level optimization to incorporate fairness into both weight and mask learning.
Fig.~\ref{fig:bifp-framework} provides an overview of BiFP.
We first initialize a mask $m_0$ and optimize the network weights with fairness constraints.
We then fix the weights and optimize the relaxed mask with a fairness constraint.
These alternating updates account for fairness at both levels of the optimization.
We \textbf{jointly optimize the mask and weight update processes} under fairness constraints.
This joint optimization balances model performance and fairness in the pruning task.

We summarize our contributions as follows.
\textbf{a)}
We study algorithmic bias in neural network pruning and compare how different pruning techniques affect bias.
\textbf{b)}
We introduce fairness notions tailored to model pruning, including fairness with respect to compression-induced performance degradation.
\textbf{c)}
We design a joint optimization framework, Bi-level Fair Pruning (BiFP), that incorporates fairness into both mask and weight learning during pruning.
This formulation avoids separate fairness and pruning optimization runs.
\textbf{d)}
We evaluate BiFP across multiple datasets, predictive tasks, architectures, and sparsity levels, and compare its fairness, accuracy, and training efficiency with existing techniques.
Ablation studies examine the respective effects of the mask and weight constraints.

\section{Related Work}

\subsection{Fair Classification}
Research on algorithmic fairness addresses two primary questions:
1) how to define and identify bias in machine learning tasks and
2) how to mitigate bias in conventional machine learning models.
Several notions formally define and measure fairness.
A widely used notion is \textit{statistical parity}, also called \textit{demographic parity}, which requires protected and non-protected groups to receive favorable decisions at similar rates.
The quantitative metrics derived from \textit{statistical parity} include \textit{risk difference}, \textit{risk ratio}, \textit{relative change}, and \textit{odds ratio}~\cite{DBLP:journals/datamine/Zliobaite17}.
Prior work also studies individual fairness, under which similar individuals should receive similar decisions~\cite{DBLP:conf/icml/ZemelWSPD13, DBLP:conf/kdd/ThanhRT11}.
Group criteria compare aggregate outcomes across populations defined by a sensitive attribute, whereas individual fairness constrains how a model treats similar instances.
We adopt a group criterion to evaluate whether pruning changes predictive accuracy unevenly across demographic groups.
Causal fairness criteria instead examine how sensitive attributes affect predictions through specified causal pathways. Wu et al.~\cite{DBLP:conf/nips/Wu0WT19} introduce path-specific counterfactual fairness and derive bounds when the criterion cannot be identified from observational data.
We refer readers to surveys such as~\cite{DBLP:journals/datamine/Zliobaite17,DBLP:journals/csur/MehrabiMSLG21} for further details.
Bias-mitigation methods are commonly categorized as pre-processing, in-processing, or post-processing.
	{Pre-processing} methods modify training data to reduce potential prejudice and discrimination before model training.
Common pre-processing methods include \textit{Massaging}~\cite{kamiran2009classifying}, \textit{Reweighting}~\cite{DBLP:conf/icdm/CaldersKP09}, and \textit{Preferential Sampling}~\cite{kamiran2012data}.
Fairness in the modified training data does not necessarily guarantee fairness in predictions.
Zhang et al.~\cite{DBLP:conf/ijcai/ZhangWW18} analyze this distinction and develop a two-phase framework with bounds on prediction-time discrimination.
	{In-processing} methods~\cite{DBLP:conf/icdm/KamiranCP10, DBLP:journals/datamine/CaldersV10, DBLP:conf/aistats/ZafarVGG17,DBLP:conf/kdd/Corbett-DaviesP17} modify learning algorithms by adding fairness constraints or regularizers to their objective functions.
	{Post-processing} methods~\cite{kamiran2012decision,dblp:conf/nips/hardtpns16,DBLP:conf/aistats/AwasthiKM20} adjust predictions produced by standard machine learning models.
Fair representation learning~\cite{DBLP:journals/corr/EdwardsS15,DBLP:conf/nips/XieDDHN17,DBLP:conf/icml/MadrasCPZ18,DBLP:conf/aies/ZhangLM18} and fair generative models~\cite{DBLP:conf/bigdataconf/XuYZW18,DBLP:journals/ibmrd/SattigeriHCV19} are also active research directions.
Edwards and Storkey~\cite{DBLP:journals/corr/EdwardsS15} learn fair representations from which adversaries cannot infer sensitive information.
Madras et al.~\cite{DBLP:conf/icml/MadrasCPZ18} develop adversarially fair representations with theoretical guarantees for group-fairness metrics.
Several studies~\cite{DBLP:conf/bigdataconf/XuYZW18,DBLP:journals/ibmrd/SattigeriHCV19} use generative adversarial networks (GANs) to create synthetic data that preserves data utility while promoting fairness across protected groups.
Xu et al.~\cite{DBLP:conf/ijcai/XuWYZW19} extend fair data generation to causal criteria by designing generators that model both the original causal graph and its intervened counterpart.

Fairness-aware learning also covers selection and imperfect data. Jiang et al.~\cite{DBLP:conf/ijcnn/JiangDW23} study fair selection through kernel density estimation, while Dai et al.~\cite{DBLP:conf/ijcnn/DaiJHZW24} use bag-level supervision to learn fair multiple-instance classifiers. FairSAM~\cite{DBLP:journals/tmlr/DaiJMW26} addresses unequal performance degradation across demographic groups under image corruption through sharpness-aware minimization. Here, we study disparities caused by model pruning.

\subsection{Neural Network Pruning}
Neural network pruning reduces the size and computational cost of deep neural networks by selectively removing weights or connections while preserving predictive performance.
Pruning techniques can be broadly categorized as unstructured or structured.
\textbf{Unstructured pruning} removes individual weights or connections without imposing a predetermined pattern.
Early methods~\cite{han2015deep,guo2016dynamic} remove weights based on their magnitudes.
Subsequent methods~\cite{baykal2018data,lin2020dynamic,yu2018nisp,molchanov2019importance} use training data to estimate the influence of each parameter.
Other approaches formulate pruning as an optimization problem~\cite{ren2019admm,zhang2018systematic,zhang2022advancing,hoefler2021sparsity,sehwag2020hydra}.
\textbf{Structured pruning} removes specific structures within the neural network, such as complete neurons, channels, or layers.
For convolutional networks, it often removes entire filters or feature maps, producing architectures that can be deployed efficiently on standard hardware.
Structured pruning has been studied extensively~\cite{liu2019metapruning,li2019learning,he2019filter,dong2017more,ye2018rethinking,ye2020good,molchanov2016pruning}.
Filter-wise pruning assigns importance scores based on weights~\cite{he2017channel,he2018soft} or data~\cite{yu2018nisp,liebenwein2019provable}.
The two pruning families produce different computational structures.
Unstructured methods provide fine-grained sparsity but may require sparse execution support.
Structured methods remove complete computational units, making their efficiency gains easier to realize on standard hardware.

\subsection{Fairness in Pruned Neural Networks}
Recent work examines how model compression, particularly neural network pruning, can introduce or amplify bias in sparse models.
Paganini et al.~\cite{paganini2020prune} investigate performance imbalances caused by pruning and propose a Pareto-based framework that incorporates fairness into the pruning process.
For text datasets, Hansen et al.~\cite{DBLP:conf/acl/HansenS21} empirically analyze fairness in the lottery-ticket extraction process.
Tran et al.~\cite{DBLP:conf/nips/0007FKN22} show that pruning can introduce bias and amplify existing model disparities.
For vision models, Iofinova et al.~\cite{DBLP:journals/corr/abs-2304-12622} analyze bias in pruned models and propose criteria for determining when pruning is likely to intensify bias.
Lin et al.~\cite{DBLP:conf/eccv/LinKJ22} introduce Fairness-aware GRAdient Pruning mEthod (FairGRAPE), which reduces disproportionate pruning effects across subgroups by greedily selecting weights that preserve relative group importance.
Tang et al.~\cite{tang_2023_cvpr} identify fair and accurate lottery tickets from randomly initialized weights.
Unlike existing approaches, BiFP couples fairness-aware mask and weight updates within a single optimization framework.

\section{Preliminaries}
\label{sec:preliminaries}

\subsection{Fair Classification}

Consider a dataset $\mathcal{D}=\big\{(\mathbf{x}_i, \mathbf{y}_i, s_i)\big\}_{i=1}^N$, where $\mathbf{x}_i \in \mathcal{X}$ is an input feature vector, $\mathbf{y}_i \in \mathcal{Y}$ is a ground-truth label, and $s_i \in \mathcal{S}$ is a sensitive attribute.
For simplicity and clarity, we present the formulation for binary labels, $\mathcal{Y}=\{-1,+1\}$.
A real-valued score function $h_\theta:\mathcal{X}\rightarrow\mathbb{R}$ is parameterized by $\theta$, and the induced classifier is $f_\theta = \operatorname{sign}(h_\theta): \mathcal{X} \rightarrow \mathcal{Y}$, with the convention $\operatorname{sign}(0)=+1$.
The classification objective is to find an optimal parameter set $\theta^*$ such that
$ \theta^* = \argmin_{\theta} \mathbb{L}(\theta),$
where $\mathbb{L}(\theta) = \frac{1}{N}\sum_{i=1}^{N} \ell_c(h_\theta(\mathbf{x}_i), \mathbf{y}_i) $ is the empirical loss and $\ell_c$ is a surrogate loss, such as hinge or logistic loss~\cite{bartlett2006convexity}.

Fairness can be incorporated into this classification task through constraints.
Specifically, we consider two demographic groups $\mathcal{D}^+=\big\{(\mathbf{x}_i, \mathbf{y}_i, s_i)|s_i=s^+\big\}$ and $\mathcal{D}^-=\big\{(\mathbf{x}_i, \mathbf{y}_i, s_i)|s_i=s^-\big\}$, corresponding to two values of the sensitive attribute, such as the male and female groups in an income-prediction task.
Given a performance metric $\mathbb{M}(f, \mathcal{D})$, such as accuracy, we say that classifier $f_\theta$ is fair if the difference between the two groups does not exceed a threshold:
\begin{equation}
	\nonumber
	\mathbb{F} (f_\theta, \mathcal{D};\mathbb{M} ) =|\mathbb{M}(f_\theta, \mathcal{D}^+) - \mathbb{M}(f_\theta, \mathcal{D}^-)|\leq \tau,
\end{equation}
where $\tau$ is the user-defined fairness threshold.
The absolute value measures the magnitude of disparity regardless of which group has higher performance.
The threshold $\tau$ specifies the tolerated disparity and separates the fairness requirement from the predictive loss minimized during training.

The metric $\mathbb{M}$ should reflect the requirements of the application domain.
As Iofinova et al.~\cite{DBLP:journals/corr/abs-2304-12622} note, measuring accuracy differences across race and gender is appropriate for facial identification as even moderate differences can produce discriminatory outcomes.
We therefore use the accuracy disparity as the fairness metric, with accuracy defined as
\begin{equation}
	\nonumber
	\mathbb{A}(f_\theta, \mathcal{D}) = \frac{1}{|\mathcal{D}|}\sum_{(\mathbf{x},\mathbf{y},s)\in\mathcal{D}}\mathds{1}(\mathbf{y}=f_\theta(\mathbf{x})),
\end{equation}
where $\mathds{1}$ is an indicator function.

The fairness-aware classification problem is then formulated as
\begin{align}
	\nonumber
	\theta^* = \argmin_{\theta} \mathbb{L}(\theta) \quad
	s.t.
	\quad \mathbb{F}(f_\theta, \mathcal{D};\mathbb{A} ) \leq \tau
\end{align}

\subsection{Neural Network Pruning}

Many neural network parameters contribute negligibly to prediction accuracy.
These redundant parameters increase the computational cost of training and inference.
Neural network pruning techniques identify and remove such parameters to obtain a more compact model.

Given a model $f$ and its score function $h$ parameterized by $\theta$, pruning seeks a sparse mask $m$ while preserving accuracy.
We formulate this objective as
\begin{align}
	\nonumber
	m^* = \argmin_{m\in\{0,1\}^d} \frac{1}{N} \sum_{i=1}^N\ell_c (h_{m \odot \theta}(\mathbf{x}_i), \mathbf{y}_i) \quad 	s.t.
	\quad ||{m}||_{0}\leq k,
\end{align}
where $k$ is the desired number of retained parameters and $d$ is the total number of prunable parameter entries.
$m\in\{0,1\}^d$ is a binary mask applied to the parameters.
$\|\cdot\|_{0}$ denotes the number of non-zero elements.
The operator $\odot$ denotes element-wise multiplication.
The loss term $\ell_c$ measures prediction loss under the selected mask.
We define mask sparsity as the proportion of zero-valued parameter entries, $1-\|m\|_0/d$.
Thus, the mask determines which entries of $\theta$ remain active, and the $\ell_0$ constraint limits the capacity of the pruned model.
This coupling between structure and prediction is central to fair pruning as removing a parameter can affect the two demographic groups differently even when aggregate accuracy changes little.

\section{Fair Neural Network Pruning}

We now integrate fairness constraints into neural network pruning.
The objective is to obtain the efficiency gains of compression without disproportionately degrading predictive performance for a demographic group.

\subsection{Fairness Notions for Compressed Models}

For a compressed model, fairness can be evaluated by comparing a selected performance metric across demographic groups.
If the difference does not exceed a specified threshold, the compressed model satisfies the corresponding fairness notion.
Let $h_c$ denote the compressed scoring function and $f_c$ denote the compressed classification model.
We define two fairness criteria as follows.

\begin{definition}[Performance Fairness]
	\label{def:performance-fairness}
	Given a compressed model $f_c$, we say that the model satisfies performance fairness if and only if its accuracy difference between two demographic groups does not exceed $\tau$, i.e.,
	$ \mathbb{F}(f_c, \mathcal{D}; \mathbb{A} ) = |\mathbb{A}(f_c, \mathcal{D}^+) - \mathbb{A}(f_c, \mathcal{D}^-)| \leq \tau.
	$
\end{definition}

In addition to the absolute group disparity of a compressed model, compression may amplify bias relative to the original model $f$.
This occurs when compression causes a larger performance reduction for one group than for the other.
To measure this effect, we define the accuracy reduction from the original model to the compressed model.

\begin{definition}[Compressed Model Performance Degradation]
	Given the original model $f$ and its compressed version $f_c$, the performance reduction is defined as $
		\mathbb{R}(f, f_c, \mathcal{D})  = \mathbb{A}(f, \mathcal{D}) - \mathbb{A}(f_c, \mathcal{D}).
	$
\end{definition}

We then define a fairness metric based on the performance degradation caused by compression.
\begin{definition}[Performance Degradation Fairness]
	\label{def:degradation-fairness}
	Given the original model $f$ and its compressed version $f_c$, we say that the compression satisfies performance degradation fairness if and only if the difference in performance reduction between two demographic groups does not exceed $\tau$, i.e.,
	$
		\mathbb{F}(f_c, \mathcal{D}; \mathbb{R}) = |\mathbb{R}(f, f_c, \mathcal{D}^+) - \mathbb{R}(f, f_c, \mathcal{D}^-)| \leq \tau.
	$
\end{definition}

\subsection{Fair and Efficient Learning via Bi-level Optimization}

Pruning applies masks to neural network weights, whereas fairness-aware learning adjusts those weights to reduce group disparities.
Applying these procedures sequentially can degrade either accuracy or fairness, as shown in Fig.~\ref{fig:pruning-fairness-motivation}.
We therefore incorporate fairness into both mask and weight learning throughout pruning.
We formulate this joint learning task as a bi-level optimization problem and derive an efficient optimization procedure.
This formulation reflects the asymmetric roles of the two variable sets.
The mask selects the sparse model structure, while the weights determine the best predictor available within that structure.
Treating these variables independently would ignore that the quality of a mask can only be assessed after the retained weights have adapted to it.

Formally, neural network pruning seeks a mask $m$ that determines the model's sparsity pattern, while fair classification seeks parameters $\theta$ that maintain accuracy subject to a fairness requirement.
For an arbitrary mask $m$ and parameter set $\theta$, we denote the masked scoring function by $h_c := h_{m \odot \theta}: \mathcal{X} \rightarrow \mathbb{R}$ and the induced compressed classifier by $f_c := f_{m \odot \theta} = \operatorname{sign}(h_{m \odot \theta})$.
We formulate fair pruning as the following bi-level optimization problem:
\begin{align}
	\label{eq:bilevel-fair-pruning}
	\min_{m\in\{0,1\}^d} \frac{1}{N}      & \sum_{i=1}^N\ell_c \big(h_{m \odot \theta^*(m)}(\mathbf{x}_i), \mathbf{y}_i \big),                     \\
	s.t.              \quad \theta^*(m) = & \argmin_{\theta}
	\frac{1}{N} \sum_{i=1}^N \ell_c \big(h_{m \odot \theta} (\mathbf{x}_i), \mathbf{y}_i \big), \nonumber                                          \\
	\qquad                                & \mathbb{F}(f_{m \odot \theta^*(m)}, \mathcal{D}; \mathbb{M}) \leq \tau,\quad \|m\|_0 \leq k. \nonumber
\end{align}

where $m$ and $\theta$ are the upper-level and lower-level optimization variables, respectively.
For a candidate mask $m$, the lower-level problem computes $\theta^*(m)$ by minimizing classification loss subject to fairness.
The upper-level problem evaluates the masked model using these adapted weights.
It therefore favors masks that admit accurate, fairness-aware lower-level solutions rather than masks that appear accurate only under the original dense weights.
The constraint $\|m\|_0\leq k$ imposes the target model size.

The accuracy-based fairness constraint in Eq.~\eqref{eq:bilevel-fair-pruning} involves discrete predictions and is not directly differentiable.
To support gradient-based optimization, we replace it with a differentiable surrogate penalty.
The relaxation must provide a trainable signal for reducing disparity and support gradient-based updates at both levels.

\begin{figure*}[h]
	\centering
	\includegraphics[width=0.8\textwidth]{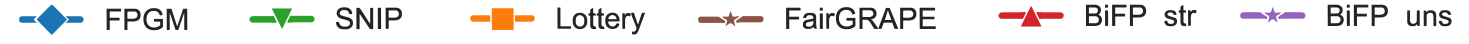}
	\vspace{-0.1em}

	\begin{subfigure}[b]{0.245\textwidth}
		\includegraphics[width=\textwidth, height=0.17\textheight]{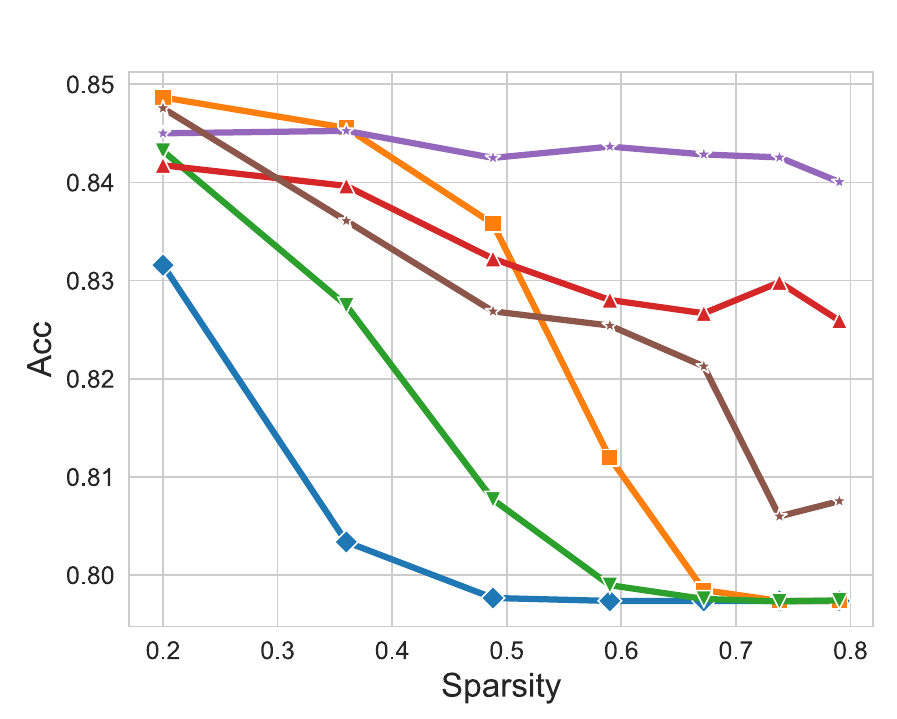}
		\caption{\small{CelebA Bags\_Under\_Eyes Acc}}
	\end{subfigure}
	\begin{subfigure}[b]{0.245\textwidth}
		\includegraphics[width=\textwidth, height=0.17\textheight]{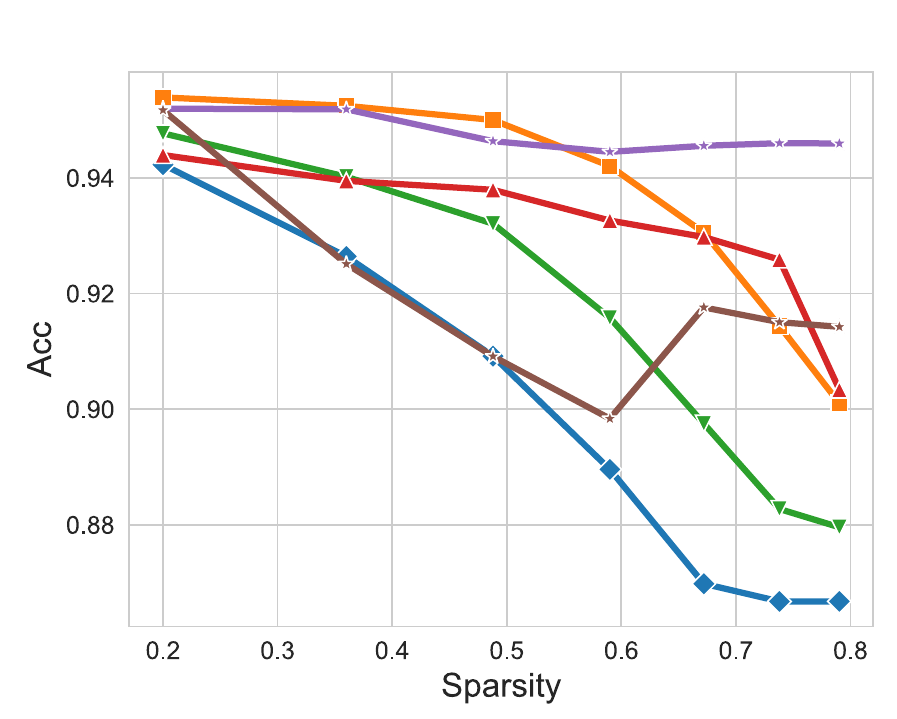}
		\caption{CelebA Blond\_Hair Acc}
	\end{subfigure}
	\begin{subfigure}[b]{0.245\textwidth}
		\includegraphics[width=\textwidth, height=0.17\textheight]{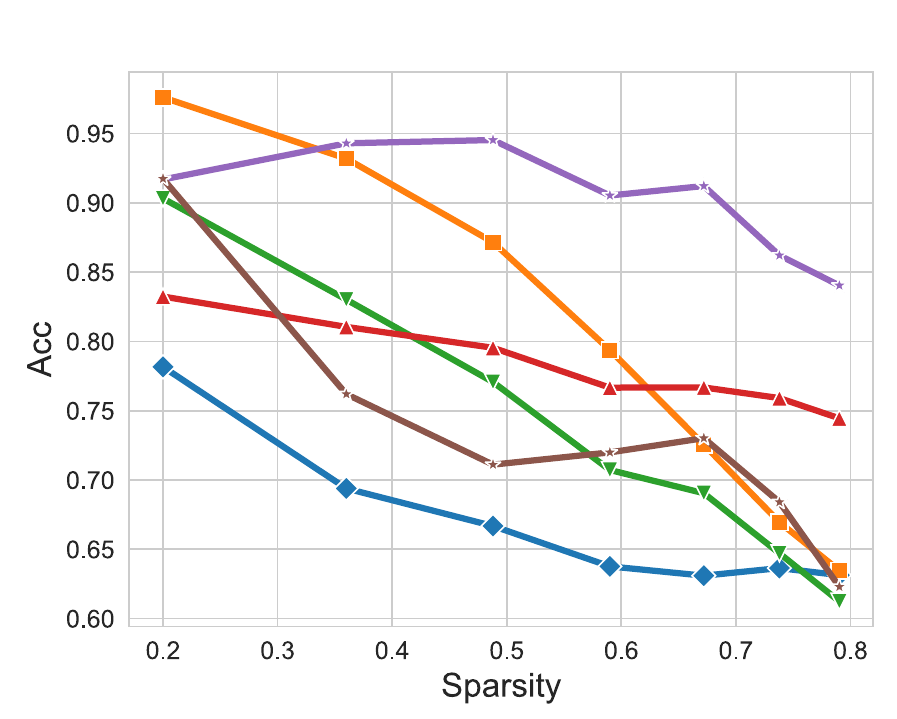}
		\caption{LFW Wavy\_Hair Acc}
	\end{subfigure}
	\begin{subfigure}[b]{0.245\textwidth}
		\includegraphics[width=\textwidth, height=0.17\textheight]{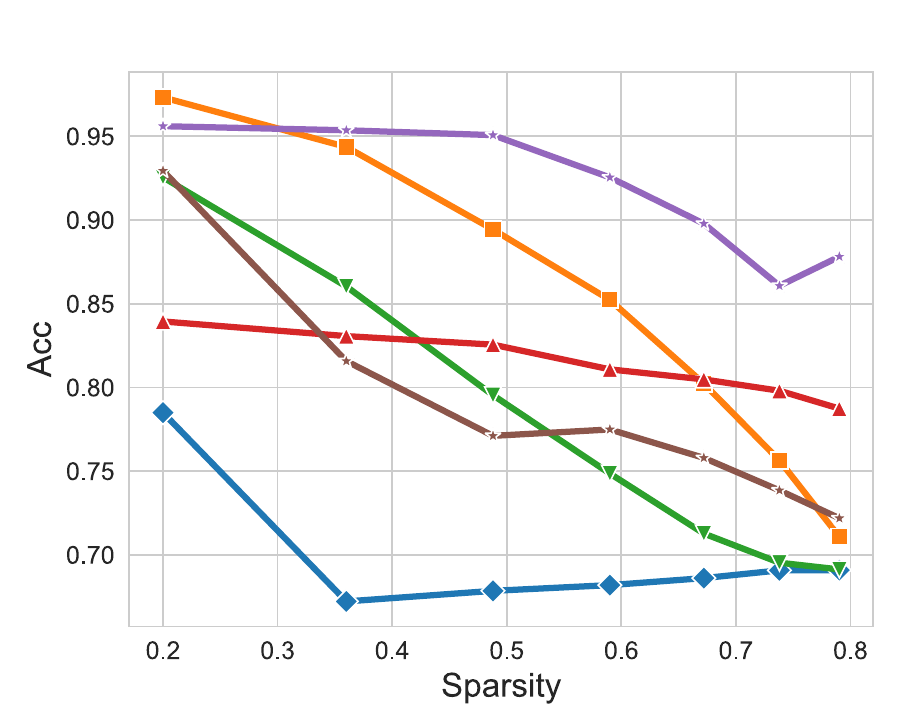}
		\caption{LFW Big\_Nose Acc}
	\end{subfigure}

	\begin{subfigure}[b]{0.245\textwidth}
		\includegraphics[width=\textwidth, height=0.17\textheight]{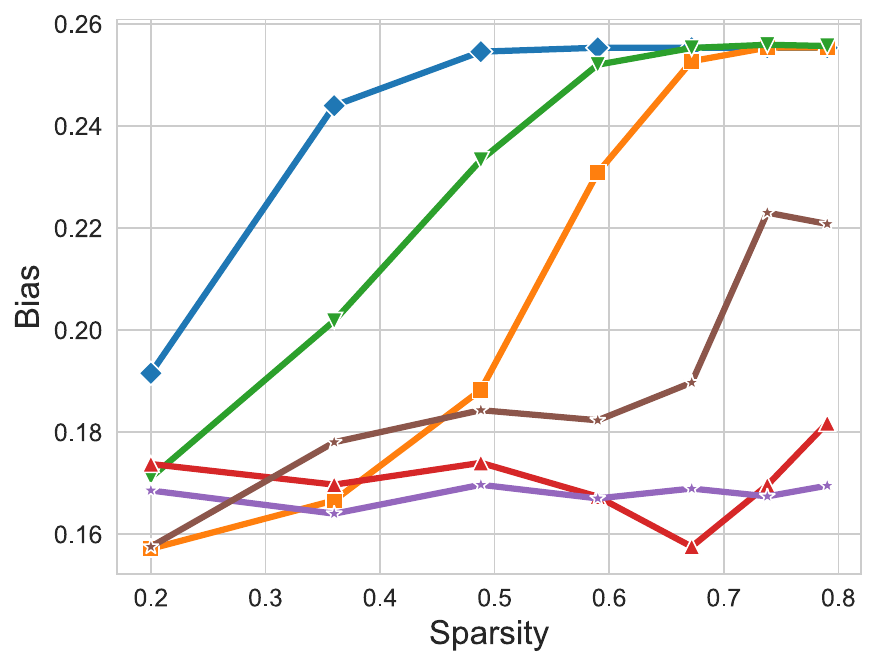}
		\caption{\small{CelebA Bags\_Under\_Eyes Bias}}
	\end{subfigure}
	\begin{subfigure}[b]{0.245\textwidth}
		\includegraphics[width=\textwidth, height=0.17\textheight]{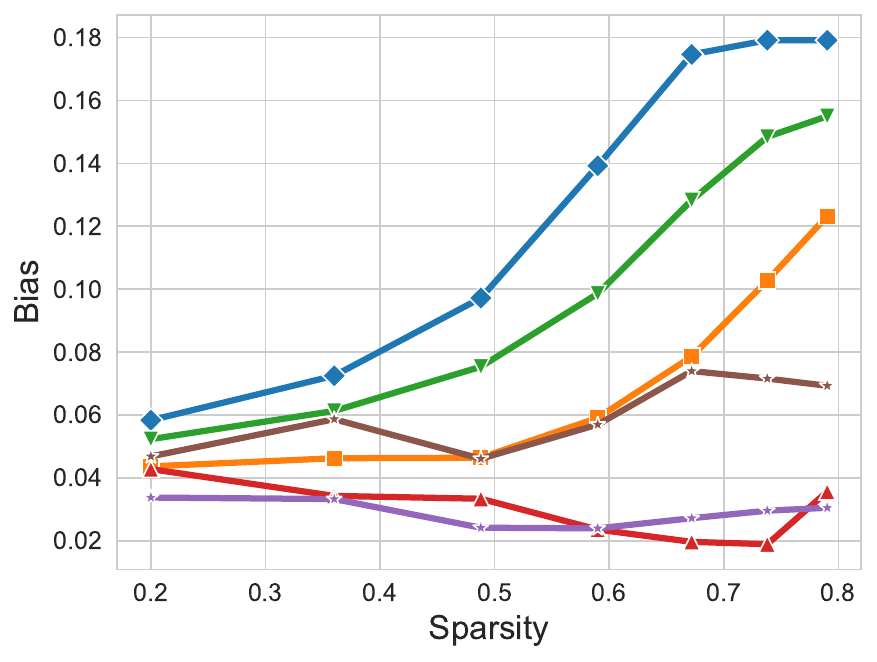}
		\caption{CelebA Blond\_Hair Bias}
	\end{subfigure}
	\begin{subfigure}[b]{0.245\textwidth}
		\includegraphics[width=\textwidth, height=0.17\textheight]{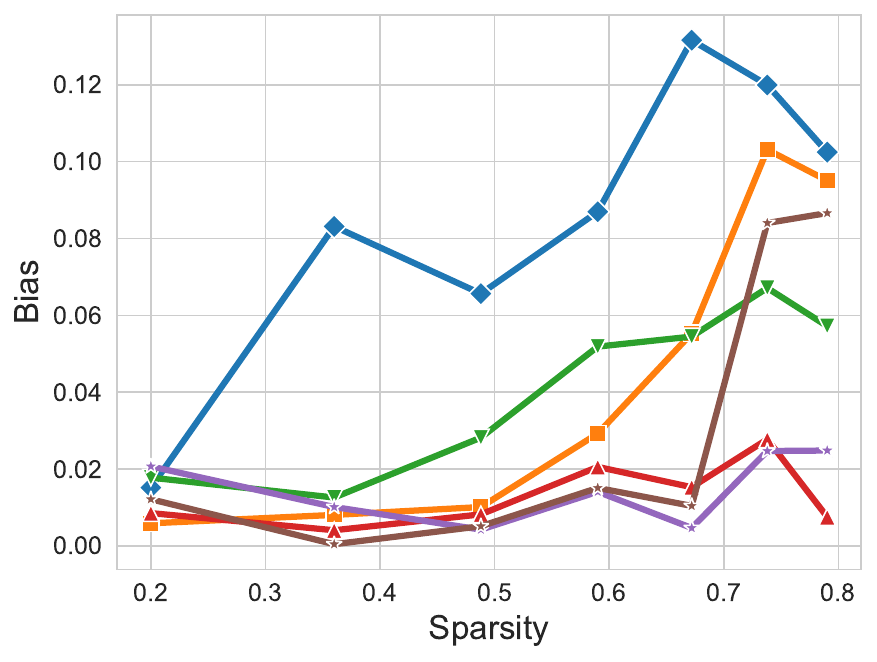}
		\caption{LFW Wavy\_Hair Bias}
	\end{subfigure}
	\begin{subfigure}[b]{0.245\textwidth}
		\includegraphics[width=\textwidth, height=0.17\textheight]{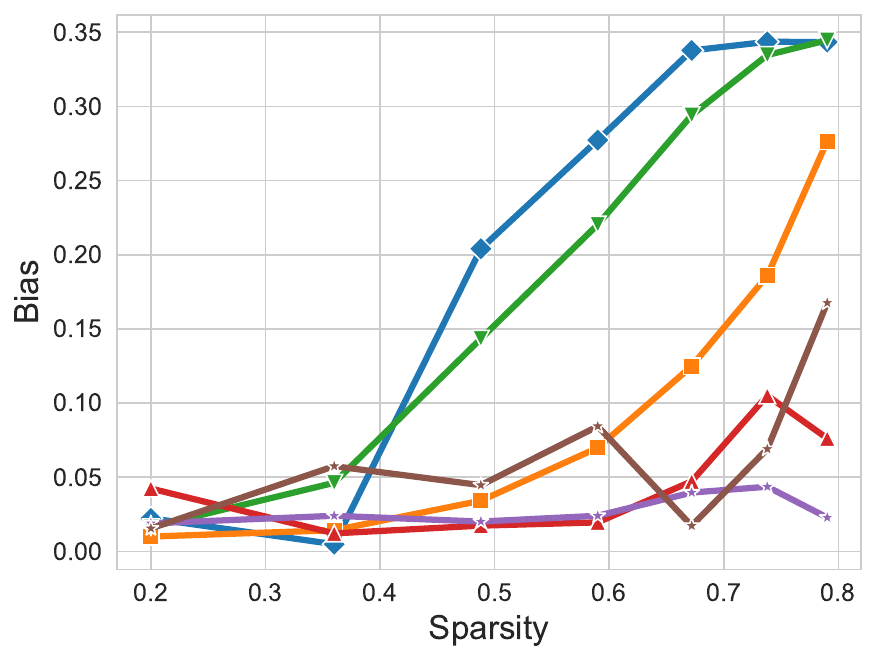}
		\caption{LFW Big\_Nose Bias}
	\end{subfigure}
	\caption{Accuracy and fairness of pruned \textbf{ResNet10} models on CelebA and LFW at varying sparsity levels.
		The prediction targets are Bags\_Under\_Eyes and Blond\_Hair for CelebA and Wavy\_Hair and Big\_Nose for LFW.
		Subfigs.~(a, b, c, d) compare the accuracy of BiFP with the baselines.
		Subfigs.~(e, f, g, h) compare the corresponding bias.
		BiFP-str and BiFP-uns maintain low bias while achieving accuracy comparable to or higher than that of the evaluated baselines.
	}
	\label{fig:resnet10-performance}
\end{figure*}

\begin{figure*}[h]
	\centering
	\includegraphics[width=0.8\textwidth]{figs/legend.pdf}
	\vspace{-0.1em}
	\begin{subfigure}[b]{0.245\textwidth}
		\includegraphics[width=\textwidth, height=0.17\textheight]{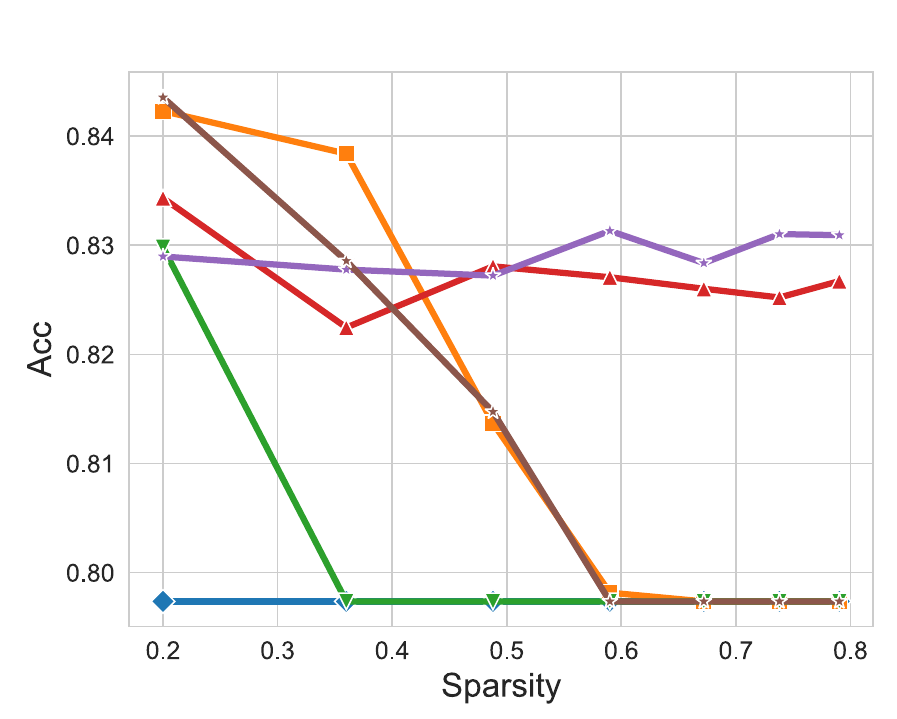}
		\caption{CelebA Bags\_Under\_Eyes Acc}
	\end{subfigure}
	\begin{subfigure}[b]{0.245\textwidth}
		\includegraphics[width=\textwidth, height=0.17\textheight]{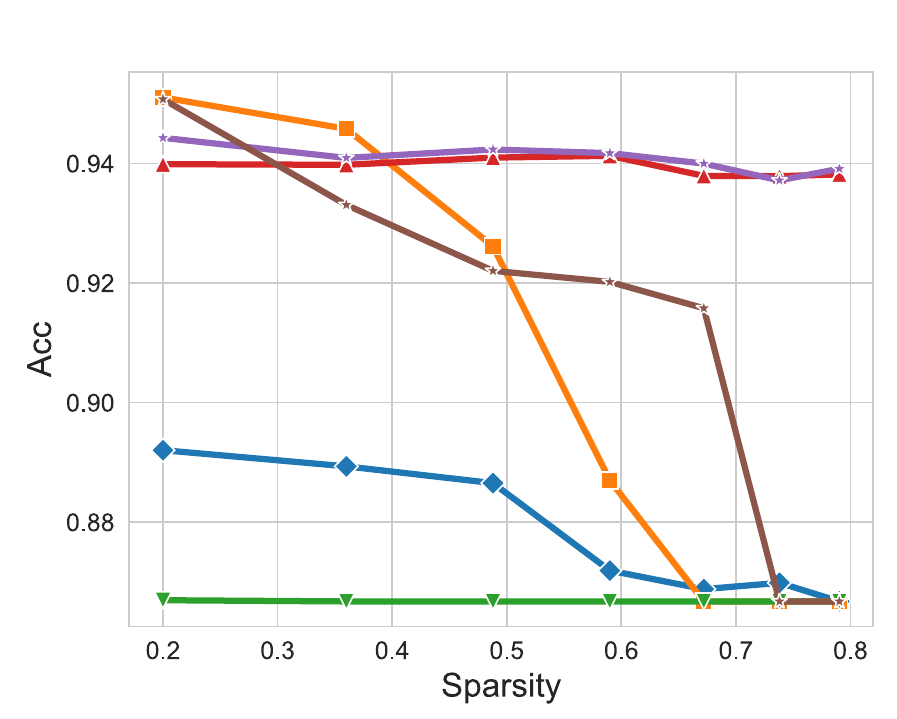}
		\caption{CelebA Blond\_Hair Acc}
	\end{subfigure}
	\begin{subfigure}[b]{0.245\textwidth}
		\includegraphics[width=\textwidth, height=0.17\textheight]{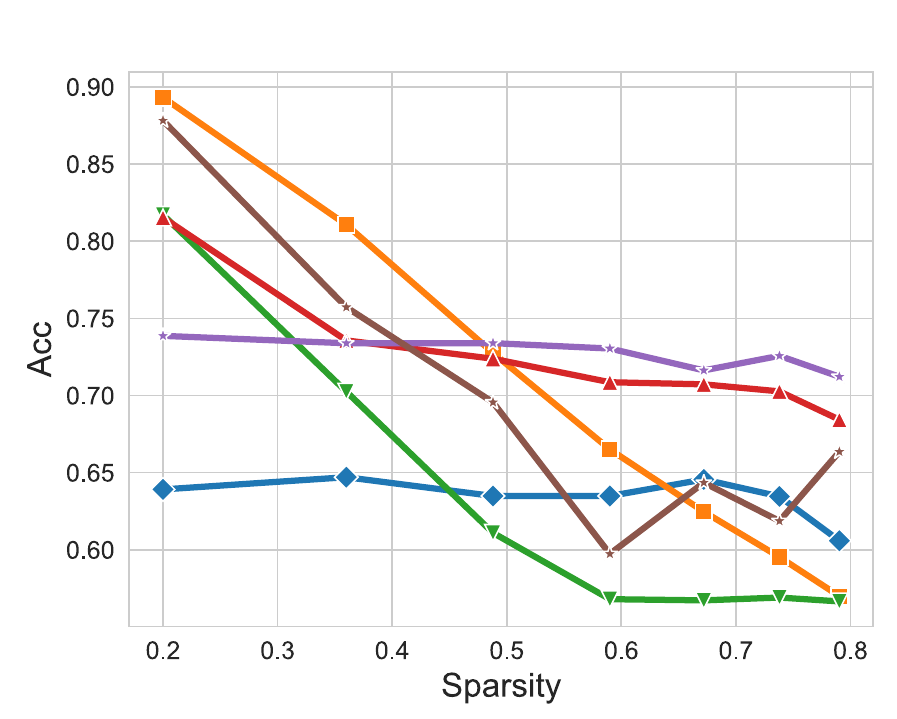}
		\caption{LFW Wavy\_Hair Acc}
	\end{subfigure}
	\begin{subfigure}[b]{0.245\textwidth}
		\includegraphics[width=\textwidth, height=0.17\textheight]{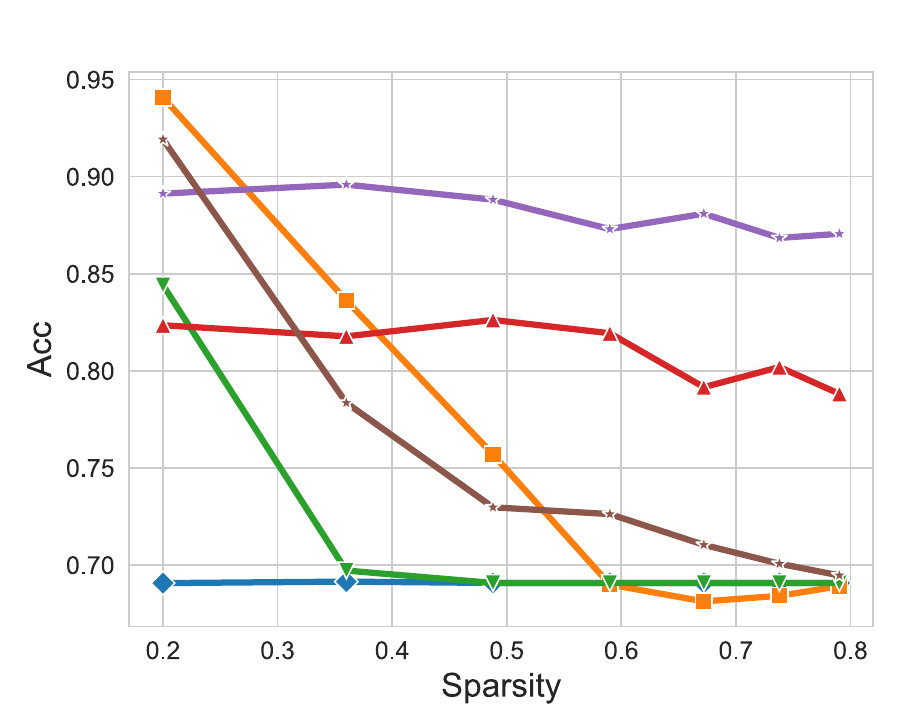}
		\caption{LFW Big\_Nose Acc}
	\end{subfigure}
	\begin{subfigure}[b]{0.245\textwidth}
		\includegraphics[width=\textwidth, height=0.17\textheight]{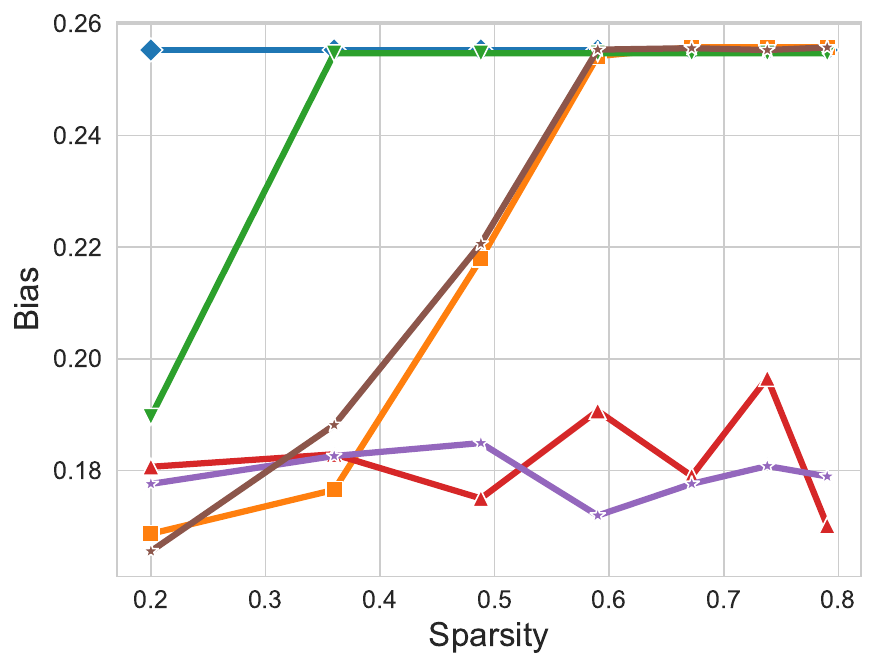}
		\caption{{CelebA Bags\_Under\_Eyes Bias}}
	\end{subfigure}
	\begin{subfigure}[b]{0.245\textwidth}
		\includegraphics[width=\textwidth, height=0.17\textheight]{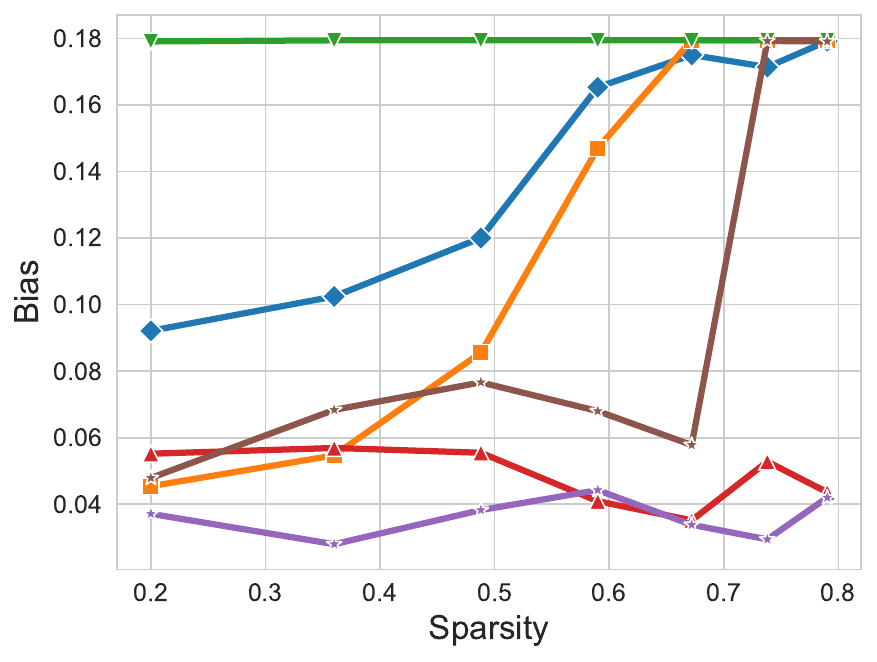}
		\caption{CelebA Blond\_Hair Bias}
	\end{subfigure}
	\begin{subfigure}[b]{0.245\textwidth}
		\includegraphics[width=\textwidth, height=0.17\textheight]{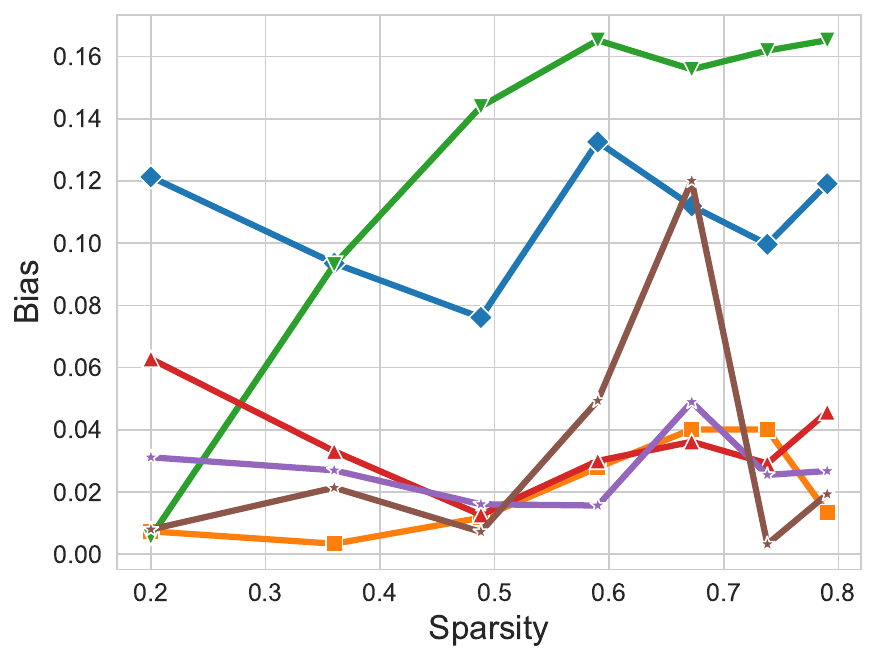}
		\caption{LFW Wavy\_Hair Bias}
	\end{subfigure}
	\begin{subfigure}[b]{0.245\textwidth}
		\includegraphics[width=\textwidth, height=0.17\textheight]{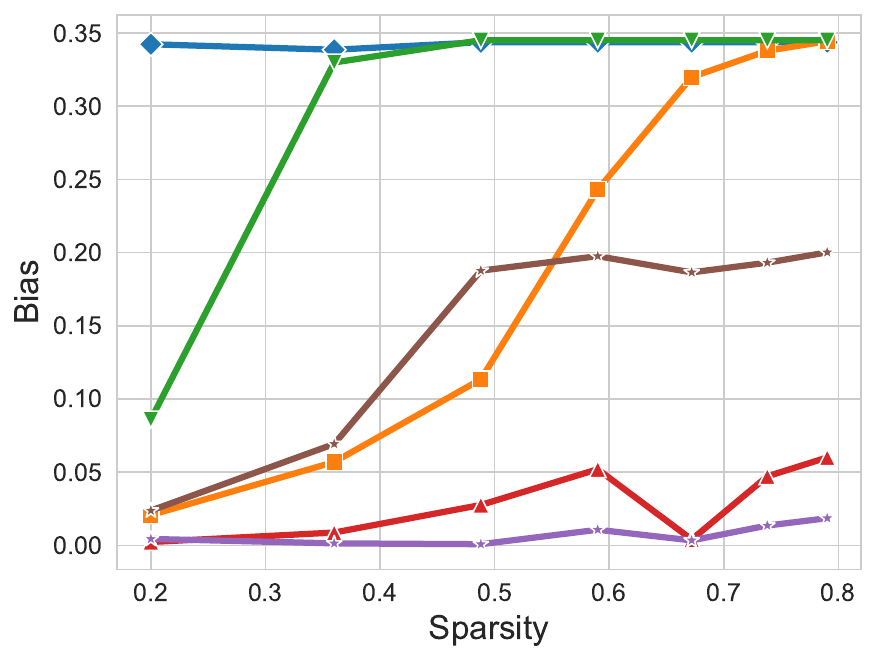}
		\caption{LFW Big\_Nose Bias}
	\end{subfigure}

	\caption{Accuracy and fairness of pruned \textbf{MobileNetV2} models on CelebA and LFW at varying sparsity levels.
		Subfigs.~(a, b, c, d) compare the accuracy of BiFP with the baselines.
		Subfigs.~(e, f, g, h) compare the corresponding bias.
		BiFP-str and BiFP-uns maintain low bias while achieving accuracy comparable to or higher than that of the evaluated baselines.
	}
	\label{fig:mobilenetv2-performance}
\end{figure*}

\subsection{Convex Relaxation of Fairness Constraint}

We express the fairness constraint in terms of accuracy.
We consider an equalized-accuracy constraint~\cite{dblp:conf/nips/hardtpns16}, under which the absolute accuracy difference between demographic groups is bounded by $\tau$.
For the differentiable relaxation below, we write the corresponding signed group-accuracy difference as
\begin{equation}
	\mathbb{F}_{\mathbb{A}} = \mathbb{E}_{X,Y|S=1}\big[\mathds{1}_{Y = f_c(X)}\big] - \mathbb{E}_{X,Y|S=-1}\big[\mathds{1}_{Y = f_c(X)}\big]
	\label{eq:signed-group-accuracy-disparity}
\end{equation}
where $\mathbb{F}_{\mathbb{A}}$ denotes the signed difference associated with $\mathbb{F}(f_{m \odot \theta}, \mathcal{D}; \mathbb{A})$ and $\mathds{1}(\cdot)$ is an indicator function.
The absolute criterion is equivalent to enforcing both $\mathbb{F}_{\mathbb{A}}\leq\tau$ and $-\mathbb{F}_{\mathbb{A}}\leq\tau$.
We derive one signed component below while the opposite direction follows by exchanging the group labels.
In the update rules, $\mathbb{F}_{\mathbb{A}}$ denotes the active signed component: the group order is chosen so that a violated component is positive, and no fairness penalty is needed when both inequalities satisfy the threshold.
To obtain a smooth and differentiable expression, we reformulate Eq.~\eqref{eq:signed-group-accuracy-disparity} as follows.
Each group-conditional expectation can be written as a reweighted expectation over the full population using the group indicator: for $P(S=1)>0$ and $P(S=-1)>0$,
\begin{align}
	\mathbb{F}_{\mathbb{A}} = & \mathbb{E}_{X,Y|S=1}\big[\mathds{1}_{Y=f_c(X)}\big] - \mathbb{E}_{X,Y|S=-1}\big[\mathds{1}_{Y=f_c(X)}\big] \nonumber \\
	=                         & \mathbb{E}_{X,Y,S}\Big[\frac{\mathds{1}_{S=1}}{P(S=1)}\mathds{1}_{Y=f_c(X)}\Big] \nonumber                           \\
	                          & - \mathbb{E}_{X,Y,S}\Big[\frac{\mathds{1}_{S=-1}}{P(S=-1)}\mathds{1}_{Y=f_c(X)}\Big] \nonumber                       \\
	=                         & \mathbb{E}_{X,Y,S}\Big[\frac{\mathds{1}_{S=1}}{P(S=1)}\mathds{1}_{Y=f_c(X)}\Big] \nonumber                           \\
	                          & + \mathbb{E}_{X,Y,S}\Big[\frac{\mathds{1}_{S=-1}}{P(S=-1)}\mathds{1}_{Y \neq f_c(X)}\Big] - 1,
	\label{eq:reweighted-accuracy-disparity}
\end{align}

When $h_c(X) \neq 0$ we have $\mathds{1}_{Y=f_c(X)} = \mathds{1}_{Y h_c(X)>0}$ for the positive group and $\mathds{1}_{Y \neq f_c(X)} = \mathds{1}_{-Y h_c(X)>0}$ for the negative group, so both terms are indicators of the single signed margin $s \cdot y \cdot h_c(x)$.
Replacing the indicators in Eq.~\ref{eq:reweighted-accuracy-disparity} with a surrogate $u(\cdot)$ and the group probabilities with the empirical frequencies $\widehat{P}(S=s) = N_s/N$ gives
\begin{equation}
	\widetilde{\mathbb{F}}_{\mathbb{A}} = \frac{1}{N}\sum_{i=1}^{N}\frac{ u \big(s_i \cdot \mathbf{y}_i \cdot h_{m \odot \theta^*(m)} ( \mathbf{x}_i ) \big)}{\widehat{P}(S=s_i)}-1.
	\label{eq:surrogate-accuracy-disparity}
\end{equation}
Thus, the surrogate $\widetilde{\mathbb{F}}_{\mathbb{A}}$ approximates $\mathbb{F}_{\mathbb{A}}$.

We consider convex surrogates $u$ that are differentiable at zero with $u'(0)>0$, such as the logistic $u(z)=\log(1+e^z)$, hinge $u(z)=\max\{0,1+z\}$, and exponential $u(z)=e^z$ surrogates~\cite{DBLP:conf/www/Wu0W19,bartlett2006convexity}.
The logistic and exponential surrogates are differentiable everywhere and the hinge surrogate uses a subgradient at $z=-1$.
For deep neural networks, the resulting objective remains nonconvex in $m$ and $\theta$, but these surrogates support gradient or subgradient updates.
The expectation form reweights prediction outcomes by the empirical group frequencies, converting the difference of group-conditional accuracies into an objective that can be estimated from training samples.
We replace the discrete indicator with $u$ so that gradient- or subgradient-based optimization is feasible.

\subsubsection{Update rule for the inner function.}
Using the differentiable surrogate for the active fairness constraint, we update the inner-level parameters by gradient descent:
\begin{equation}
	\theta^t = \theta^{t-1} - \alpha \cdot \nabla_{\theta}(\frac{1}{N}\sum_{i=1}^N \ell_c + \widetilde{\mathbb{F}}_{\mathbb{A}}) \nonumber
\end{equation}

\subsubsection{Update rule for the upper function.}
In the upper-level problem, the mask is binary and discrete.
Following~\cite{DBLP:conf/cvpr/RamanujanWKFR20}, we relax the binary mask to continuous scores $m \in [0, 1]$.
The gradient update for $m$ accounts for both its direct effect and its indirect effect through $\theta^*(m)$:
\begin{align}
	\operatorname{grad}(m) = & \nabla_m \left[\ell\big(m \odot \theta^*(m)\big) + \widetilde{\mathbb{F}}_{\mathbb{A}}\right]                                           \\
	                         & +\frac{d\theta^*(m)}{dm}\nabla_{\theta} \left[\ell\big(m \odot \theta^*(m)\big) + \widetilde{\mathbb{F}}_{\mathbb{A}}\right], \nonumber
\end{align}
where $\nabla_m$ and $\nabla_\theta$ denote partial derivatives with respect to the two arguments of the bivariate objective.
We then update $m^{t}$ as
\begin{equation}
	m^t = m^{t-1} - \beta \cdot \operatorname{grad}(m) \nonumber
\end{equation}

\section{Experiments}
\label{sec:experiments}

\definecolor{cadetgrey}{rgb}{0.57, 0.64, 0.69}
\definecolor{sliver}{rgb}{0.95,0.95, 0.95}
\colorlet{myred}{red!30}
\colorlet{mygrey}{cadetgrey!30}
\newcommand{\hlred}[1]{{\sethlcolor{myred}\hl{#1}}}
\newcommand{\hlgrey}[1]{{\sethlcolor{mygrey}\hl{#1}}}

\begin{figure*}[h]
	\centering
	\includegraphics[width=1.\textwidth]{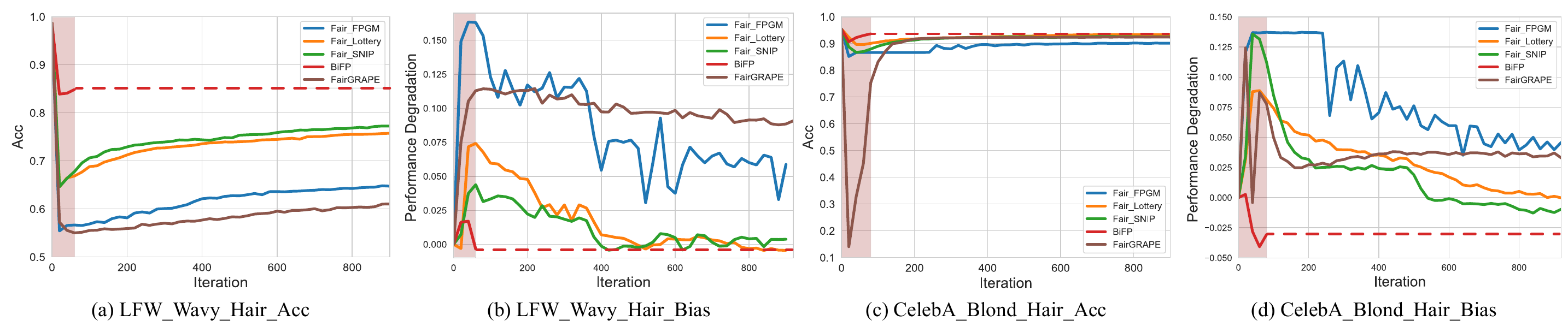}
	\caption{Training iterations required to obtain fair models with the two-stage baseline strategy.
		The \textcolor{red}{\hlred{red region}} indicates the iterations required by \textbf{BiFP}.
		The dashed line extends the final performance of \textbf{BiFP} to facilitate comparison with the baselines.
	}
	\label{fig:training-efficiency}
\end{figure*}

\noindent\textbf{Datasets and Experimental Settings.}
We evaluate the methods on two real-world image datasets with multiple prediction tasks: CelebA (202,599 face images)~\cite{liu2015faceattributes} and LFW (13,233 images)~\cite{LFWTech}.
We follow configurations used in related work (e.g., Tang et al.~\cite{tang_2023_cvpr}).
We use the original training and test splits provided by CelebA and LFW to maintain consistency with prior work.
The sensitive attribute is \texttt{Gender} for both datasets.
For CelebA, we select \texttt{Blond\_Hair} and \texttt{Bags\_Under\_Eyes} as target attributes.
For LFW, we select \texttt{Wavy\_Hair} and \texttt{Big\_Nose} as target attributes.
We pretrain ResNet10~\cite{DBLP:conf/cvpr/HeZRS16} and MobileNetV2~\cite{sandler2018mobilenetv2} networks on the training set, prune and fine-tune them on the validation set, and report test accuracy and fairness violations at different sparsity levels.
	{We run each experiment five times and report the average performance for each method.
		We train using SGD with a learning rate of $0.001$, weight decay of $5\times10^{-4}$, and momentum of $0.9$.
	}
We follow the sparsity-ratio setting of the {Lottery Ticket Hypothesis} (LTH)~\cite{frankle2018lottery} to ensure a consistent comparison.
	{All models are implemented in PyTorch and evaluated on a Linux server with an Intel Core i9-10900X CPU and an NVIDIA GeForce RTX 3070 GPU.}
All the code and data are available at \url{http://tiny.cc/fair-pruning}.
The tasks provide multiple combinations of dataset, target attribute, architecture, and sparsity level under a common sensitive attribute.

\noindent\textbf{Methods.}
We compare against the following baseline groups: unstructured pruning, structured pruning, and fairness-aware pruning.
The unstructured methods are \textbf{Single-shot Network Pruning} (\textbf{SNIP})~\cite{lee2018snip}, which computes connection sensitivity and prunes low-sensitivity edges, and \textbf{The Lottery Ticket Hypothesis} (\textbf{Lottery})~\cite{frankle2018lottery}, which identifies a sparse subnetwork that can achieve test accuracy similar to that of the original dense network.
The structured pruning baseline is \textbf{Filter Pruning via Geometric Median} (\textbf{FPGM})~\cite{he2019filter}, which prunes filters according to their geometric-median distances.
\textbf{Fairness-aware GRAdient Pruning mEthod} (\textbf{FairGRAPE})~\cite{DBLP:conf/eccv/LinKJ22} computes weight importance for each subgroup and greedily selects weights to preserve relative importance across groups.
The proposed \textbf{Bi-level Fair Pruning} (\textbf{BiFP}) supports both structured and unstructured pruning.
We use \textbf{BiFP-str} and \textbf{BiFP-uns} to denote structured and unstructured variants, respectively.
SNIP and Lottery are unstructured pruning strategies that do not optimize the fairness objective, while FPGM is a structured counterpart.
FairGRAPE incorporates subgroup information into weight selection and therefore tests whether joint mask-weight optimization offers benefits beyond fairness-aware pruning alone.
The two BiFP variants use the same general fairness formulation with different mask structures.

\noindent\textbf{Performance and Fairness.}
We first evaluate whether \textbf{BiFP} reduces fairness disparity during pruning.
Figs.~\ref{fig:resnet10-performance} and \ref{fig:mobilenetv2-performance} compare the accuracy and fairness disparities of several methods on CelebA and LFW using \textbf{ResNet10} and \textbf{MobileNetV2} backbones.
The first row of Fig.~\ref{fig:resnet10-performance} shows that \textbf{BiFP-str} and \textbf{BiFP-uns} achieve accuracy comparable to that of conventional pruning methods on ResNet10.
The second row shows that the fairness disparities of conventional methods, including \textbf{FPGM}, \textbf{Lottery}, and \textbf{SNIP}, generally increase with sparsity, whereas \textbf{BiFP-str} and \textbf{BiFP-uns} maintain low disparities across the evaluated sparsity levels.
For CelebA with \texttt{Blond\_Hair} as the target attribute and \texttt{Gender} as the sensitive attribute, BiFP achieves $6\%$ higher accuracy than the baselines at $80\%$ sparsity while also reducing the fairness disparity.
	{\textbf{FairGRAPE} preserves fairness at low sparsity levels but exhibits larger disparities at high sparsity levels.}
With MobileNetV2, \textbf{BiFP-str} and \textbf{BiFP-uns} achieve accuracy comparable to that of the baselines at low sparsity.
When sparsity exceeds $0.5$, BiFP obtains higher accuracy than the evaluated baselines.
BiFP also yields lower fairness disparities across the evaluated sparsity levels.
These results indicate that BiFP is particularly effective under high sparsity.
When most parameters remain active, the methods retain similar capacity and their accuracy values are close.
As pruning becomes more aggressive, mask selection has a larger effect on which predictive features remain available.
The consistent trends across ResNet10 and MobileNetV2 show that the advantage of BiFP holds across different backbone architectures.

\begin{figure}[h]
	\centering
	\includegraphics[width=\linewidth]{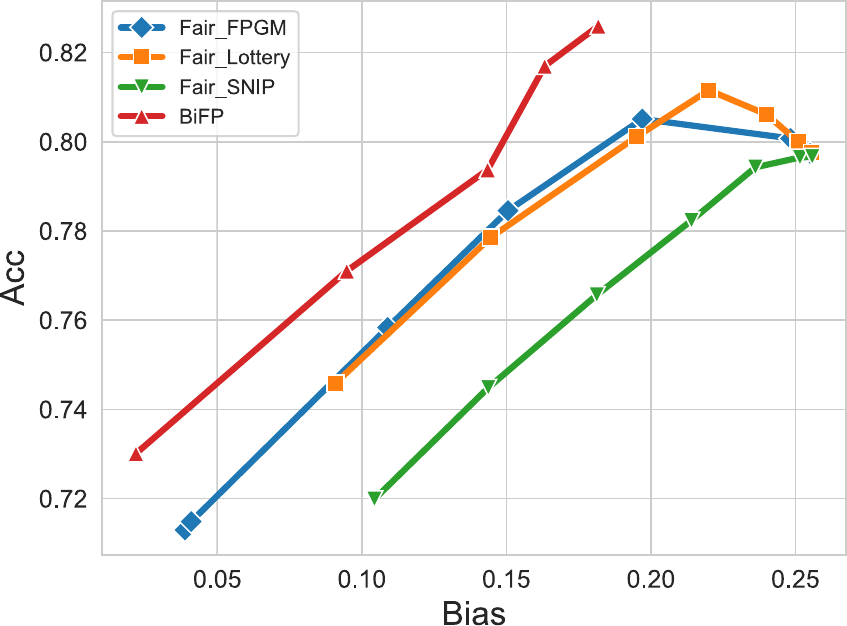}
	\caption{Accuracy-fairness trade-off for Fair\_FPGM, Fair\_Lottery, Fair\_SNIP, and \textbf{BiFP}.
		Accuracy (Acc) is plotted against bias.
		BiFP attains lower bias at a given accuracy and higher accuracy at a given bias than the evaluated baselines.
	}
	\label{fig:accuracy-fairness-tradeoff}
\end{figure}

\noindent\textbf{Trade-off Between Accuracy and Fairness.}
In the remaining experiments, we focus on the structured variant of \textbf{BiFP} and omit the suffix for simplicity.
We investigate its accuracy-fairness trade-off on CelebA with \texttt{Blond\_Hair} as the target attribute at $70\%$ sparsity. Fig.~\ref{fig:accuracy-fairness-tradeoff} reports the results.
The curves show that BiFP attains a lower fairness disparity than the baselines at a given accuracy and higher accuracy at a given disparity.

\noindent\textbf{Fairness and Training Efficiency.}
To compare fairness and training efficiency, we apply a two-stage fair training strategy to the baselines.
In the first stage, we prune a dense model using a conventional pruning technique.
In the second stage, we retrain the sparse model with a fairness constraint to achieve fairness.
	{For \textbf{FairGRAPE}, we follow the original implementation and train until performance stabilizes.}
We record the number of training iterations for BiFP and the baselines in Fig.~\ref{fig:training-efficiency}.
The red region indicates the iterations used for \textbf{BiFP} to achieve the desired fairness and accuracy.
The baselines require more iterations to reach comparable accuracy and fairness.
BiFP matches or exceeds the accuracy and fairness of the best baselines while requiring $94.22\%$ and $94.05\%$ fewer training iterations on LFW and CelebA, respectively.
The iteration comparison reflects optimization effort under the reported training procedures.
In particular, the two-stage baselines first construct a sparse model and then require additional updates to satisfy the fairness objective, whereas BiFP incorporates that objective while learning the sparse model.
We use training iterations as a common algorithmic measure, which does not account for possible differences in per-iteration runtime.

\begin{figure*}[!tbh]
	\centering
	\includegraphics[width=0.9\textwidth]{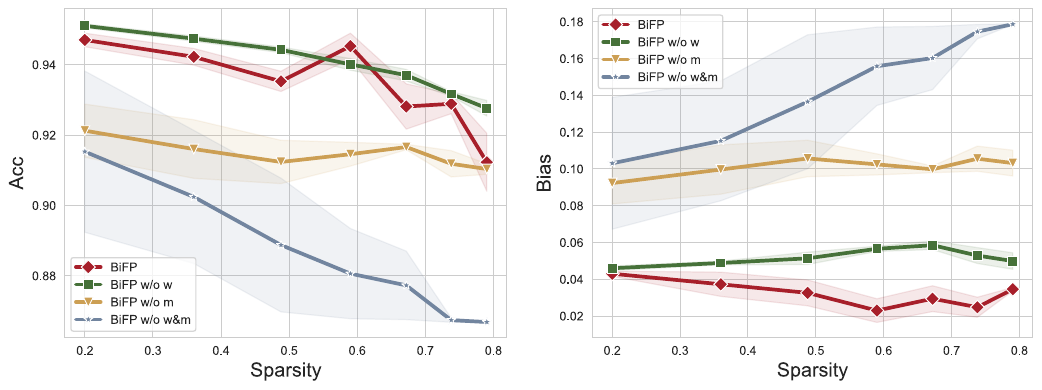}
	\caption{Ablation study of \textbf{BiFP} across varying sparsity levels on CelebA with \texttt{Blond\_Hair} as the target attribute, using \textbf{ResNet10}.
		The left and right plots report accuracy and bias, respectively, for the full method and variants without specific constraints.
		Shaded regions indicate standard deviation.
	}
	\label{fig:constraint-ablation}
\end{figure*}

\begin{figure*}[!tbh]
	\centering
	\includegraphics[width=0.9\textwidth]{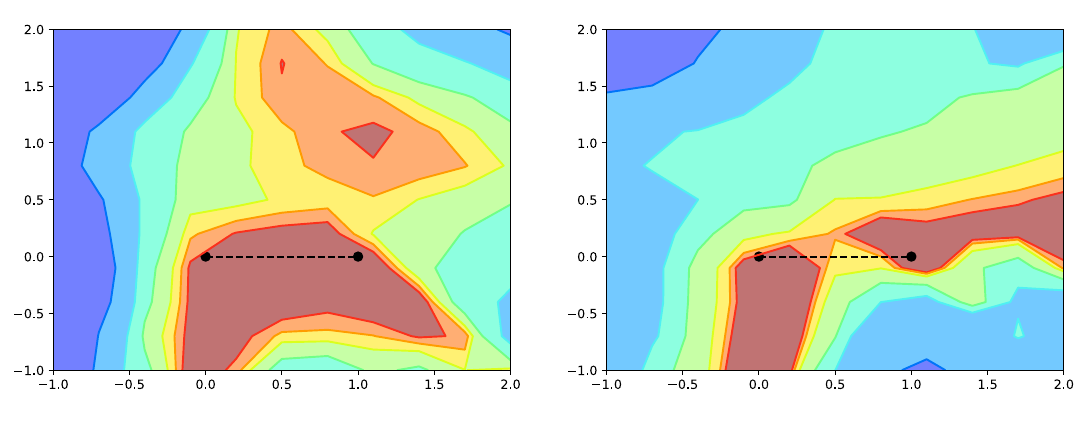}
	\caption{Loss-surface interpolation from \textbf{BiFP} to its variants: BiFP without weight constraints (\textbf{BiFP w/o w}, left) and BiFP without mask constraints (\textbf{BiFP w/o m}, right).
		The plots show how removing each constraint changes the loss along the interpolation path.
	}
	\label{fig:loss-interpolation}
\end{figure*}

\noindent\textbf{Ablation Studies.}
BiFP applies fairness constraints to both weights and masks during pruning.
We conduct an ablation study to examine the effects of these two components separately on CelebA with \texttt{Blond\_Hair} as the target attribute and \texttt{Gender} as the sensitive attribute, using the \textbf{ResNet10} backbone.
We remove the fairness penalty from the mask update and denote this variant by \textbf{BiFP w/o m}.
Similarly, we remove the fairness penalty from the weight update and denote the resulting variant by \textbf{BiFP w/o w}.
We also remove both penalties, yielding \textbf{BiFP w/o w\&m}.
Fig.~\ref{fig:constraint-ablation} presents the results.
\textbf{BiFP w/o m} and \textbf{BiFP w/o w\&m} have lower accuracy and higher fairness disparity than the full \textbf{BiFP}, indicating that fairness-aware mask learning contributes to both objectives.
\textbf{BiFP} attains accuracy comparable to that of \textbf{BiFP w/o w} while achieving the lowest disparity at all evaluated sparsity levels.
Without the weight-level fairness constraint, \textbf{BiFP w/o w} relies on the fairness-aware mask alone.
As sparsity increases, mask optimization alone is insufficient to preserve both accuracy and fairness.
Consequently, \textbf{BiFP w/o w} can retain relatively high accuracy while exhibiting a larger fairness disparity as its weights are not jointly optimized for fairness under the learned mask.

\noindent\textbf{Loss Surface Exploration.}
We further compare the loss surfaces of \textbf{BiFP} and its ablated variants.
Fig.~\ref{fig:loss-interpolation} shows linear interpolation paths from \textbf{BiFP} to \textbf{BiFP w/o w} (left) and from \textbf{BiFP} to \textbf{BiFP w/o m} (right).
The right plot shows a high-loss barrier along the path between \textbf{BiFP} and \textbf{BiFP w/o m}, indicating that the two endpoints occupy distinct regions of the loss landscape.
In contrast, the left plot shows a comparatively low-loss path between \textbf{BiFP} and \textbf{BiFP w/o w}.
These interpolation results are consistent with the ablation results.

\section{Conclusions and Future Work}

This paper examined how neural networks can be pruned without disproportionately reducing accuracy for particular demographic groups.
We proposed Bi-level Fair Pruning (\textbf{BiFP}), which learns the pruning mask and the model weights under fairness constraints.
Unlike sequential approaches that address fairness before or after pruning, BiFP accounts for fairness while determining both which parameters to retain and how those parameters should be updated.
Experiments on CelebA and LFW using ResNet10 and MobileNetV2 show that the structured and unstructured versions of BiFP maintain competitive accuracy and generally lower group disparity than the baseline pruning methods.
Regarding efficiency, BiFP also requires $94.22\%$ and $94.05\%$ fewer training iterations than the two-stage baselines on LFW and CelebA, respectively, while achieving comparable or better accuracy and fairness.
The ablation results show that both the weight and mask components are needed.
Removing the mask constraint reduces accuracy and fairness, while optimizing only a fairness-aware mask is not sufficient at high sparsity.
These results suggest that fairness needs to be considered during pruning rather than only during subsequent retraining.

Our experiments cover two face datasets, binary sensitive groups, four prediction tasks, and relatively small architectures.
Training iterations provide only an algorithmic measure of efficiency and may not reflect wall-clock cost.
In addition, the current evaluation does not consider intersectional disparities, and convergence of the relaxed bi-level procedure remains to be studied.
Further work could test BiFP on larger models and datasets from other domains, measure end-to-end computational cost, establish theoretical properties of the optimization procedure, and consider multi-group, intersectional, and other fairness criteria.

\section*{Acknowledgments}
This work was supported in part by NSF 2520496.

\bibliographystyle{IEEEtran}

\end{document}